%% file: wsdm.tex
\documentclass[sigconf]{acmart}
\renewcommand\footnotetextcopyrightpermission[1]{} % removes footnote with conference information in first column
\usepackage{booktabs} % For formal tables
\usepackage{listings}
\usepackage{color}
\usepackage[skip=2pt]{caption}
\usepackage{subcaption}
\usepackage{cleveref}
\usepackage{textcomp}

\setcopyright{none}

\DeclareCaptionFormat{captionformat}{\fontsize{7}{8}\selectfont#1#2#3}
\definecolor{dkgreen}{rgb}{0,0.6,0}
\definecolor{gray}{rgb}{0.5,0.5,0.5}
\definecolor{mauve}{rgb}{0.58,0,0.82}

\begin{document}
\title{Applying Deep Learning To Airbnb Search}
\author{Malay Haldar, Mustafa Abdool, Prashant Ramanathan, Tao Xu, Shulin Yang, Huizhong Duan, Qing Zhang, Nick Barrow-Williams, Bradley C. Turnbull, Brendan M. Collins and Thomas Legrand}
\affiliation{%
  \institution{Airbnb Inc.}
}
\email{malay.haldar@airbnb.com}

\renewcommand{\shortauthors}{Malay Haldar et al.}

\begin{abstract}
The application to search ranking is one of the biggest machine learning success stories at Airbnb. Much of the initial gains were driven by a gradient boosted decision tree model. The gains, however, plateaued over time. This paper discusses the work done in applying neural networks in an attempt to break out of that plateau. We present our perspective not with the intention of pushing the frontier of new modeling techniques. Instead, ours is a story of the elements we found useful in applying neural networks to a real life product. Deep learning was steep learning for us. To other teams embarking on similar journeys, we hope an account of our struggles and triumphs will provide some useful pointers. Bon voyage!
\end{abstract}

\begin{CCSXML}
<ccs2012>
 <concept>
  <concept_desc>Information systems~Information retrieval~Retrieval models and ranking~Learning to rank</concept_desc>
  <concept_significance>500</concept_significance>
 </concept>
 <concept>
  <concept_desc>Computing methodologies~Machine learning~Machine learning approaches~Neural networks</concept_desc>
  <concept_significance>500</concept_significance>
 </concept>
 <concept>
  <concept_desc>Applied computing~Electronic commerce~Online shopping</concept_desc>
  <concept_significance>300</concept_significance>
 </concept>
</ccs2012>
\end{CCSXML}

\ccsdesc[500]{Retrieval models and ranking~Learning to rank}
\ccsdesc[500]{Machine learning approaches~Neural networks}
\ccsdesc[300]{Electronic commerce~Online shopping}

\keywords{Search ranking, Deep learning, e-commerce}

\maketitle

\input{wsdmbody-conf}

%\bibliographystyle{ACM-Reference-Format}
\bibliography{wsdm-bibliography}

\end{document}

%% file: wsdmbody-conf.tex
\section{Introduction}
The home sharing platform at Airbnb is a two sided marketplace for hosts to rent out their spaces, referred to as listings, to be booked by prospective guests from all around the world. A typical booking starts with the guest issuing a search at airbnb.com for homes available in a particular geographic location. The task of search ranking is to respond to the guest with an ordered list of a handful of listings from the thousands available in the inventory.
 
The very first implementation of search ranking was a manually crafted scoring function. Replacing the manual scoring function with a gradient boosted decision tree (GBDT) model gave one of the largest step improvements in homes bookings in Airbnb's history, with many successful iterations to follow. The gains in online bookings eventually saturated with long spells of neutral experiments. This made the moment ripe for trying sweeping changes to the system.
 
Starting with this background, the current paper discusses our experiences in transitioning one of the at-scale search engines on the internet to deep learning. The paper is targeted towards teams that have a machine learning system in place and are starting to think about neural networks (NNs). For teams starting to explore machine learning, we would recommend a look at \cite{rulesofml} as well.

The search ranking model under discussion is part of an ecosystem of models, all of which contribute towards deciding which listings to present to the guest. These include models that predict the likelihood of a host accepting the guest's request for booking, models that predict the probability the guest will rate the on trip experience 5-star, etc. Our current discussion is focused on one particular model in this ecosystem. Considered the most complex piece of the ranking puzzle, this model is responsible for ordering the listings available according to the guest's likelihood of booking.

\begin{figure}
\includegraphics[height=1.2in, width=3.2in]{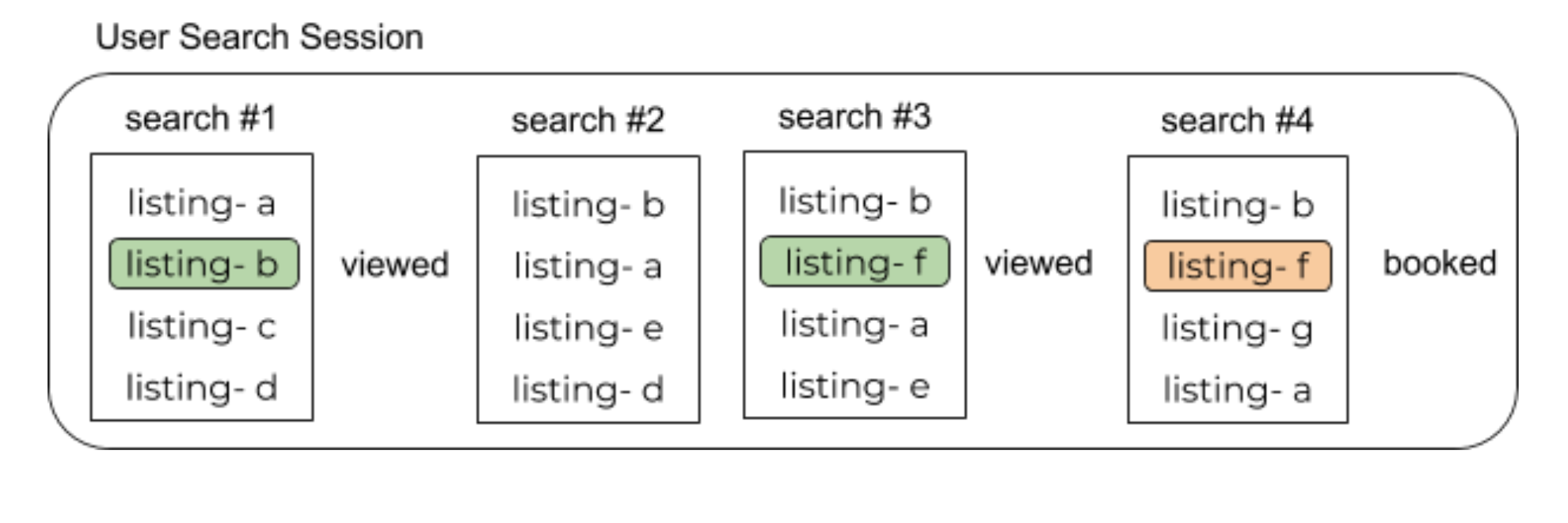}
\caption{Example search session}
\label{fig:searchsession}
\end{figure}

A typical guest search session is depicted in Figure~\ref{fig:searchsession}. It is common for guests to do multiple searches, clicking through some of the listings to view their details page. Successful sessions end with the guest booking a listing. The searches performed by a guest and their interactions are logged. While training, a new model has access to the logs of the current and previous models used in production. The new model is trained to learn a scoring function that assigns impressions of the booked listings in the logs at as high a rank as possible, similar to ~\cite{amazon}. The new model is then tested online in an A/B testing framework to see if it can achieve a statistically significant increase in conversions compared to the current model.

An overview of the paper: we start off with a summary of how the model architecture evolved over time. This is followed by feature engineering and system engineering considerations. We then describe some of our tooling and hyper-parameter explorations, finishing with a retrospective. 

\section{Model Evolution}
\begin{figure}[htp]
  \begin{subfigure}[b]{0.4\textwidth}
     \includegraphics[height=2in, width=3.2in]{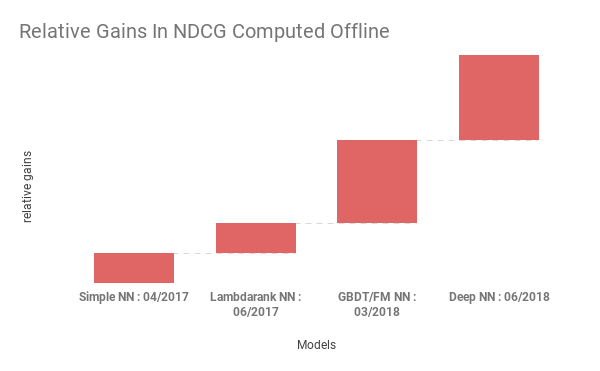}
     \label{fig:ndcggains}
     \caption{NDCG gains offline}
  \end{subfigure}
   \begin{subfigure}[b]{0.4\textwidth}
      \includegraphics[height=2in, width=3.2in]{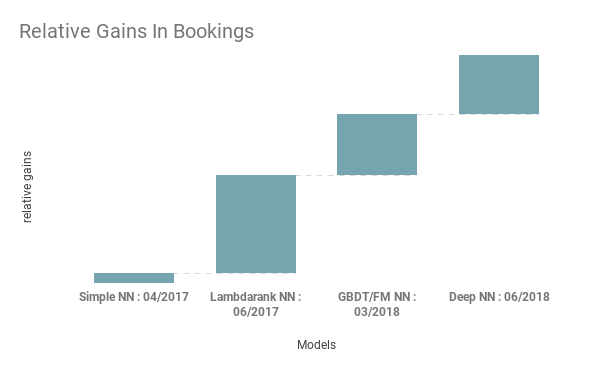}
      \label{fig:bookinggains}
      \caption{Booking gains online}
   \end{subfigure}
   \caption{Relative gains across models}
   \label{fig:modelevolution}
\end{figure}
Our transition to deep learning wasn't the result of an atomic move; it was the culmination of a series of iterative refinements. Figure \ref{fig:modelevolution}a shows the comparative improvements in our principal offline metric NDCG (normalized discounted cumulative gain), where the impression of the booked listing is assigned a relevance of 1, and all other impressions 0 relevance. The x-axis depicts the models along with the time they were launched to production.

Figure \ref{fig:modelevolution}b shows the comparative increase in conversions from the models. Overall, this represents one of the most impactful applications of machine learning at Airbnb. The sections to follow briefly describe each of these models.

\subsection{Simple NN}
Andrej Karpathy has advice regarding model architecture: don't be a hero \cite{dontbeahero}. Well, that's not how we started. Driven by "Why can't we be heroes?", we started off with some intricate custom architectures, only to get overwhelmed by their complexity and ended up consuming cycles.

The first architecture that we finally managed to get online was a simple single hidden layer NN with $32$ fully connected ReLU activations that proved booking neutral against the GBDT model. The NN was fed by the same features as the GBDT model. The training objective for the NN was also kept invariant w.r.t the GBDT model: minimizing the L2 regression loss where booked listings are assigned a utility of $1.0$ and listings that are not booked a utility of $0.0$.

The value of the whole exercise was that it validated that the entire NN pipeline was production ready and capable of serving live traffic. Aspects of this pipeline are discussed later under the feature engineering and system engineering sections.

\subsection{Lambdarank NN}
Not being a hero got us off to a start, but not very far. In time we would adapt Karpathy's advice to: don't be a hero, in the beginning. Our first breakthrough came when we combined a NN with the idea behind Lamdarank~\cite{lambdarank}. Offline we were using NDCG as our principal metric. Lambdarank gave us a way to directly optimize the NN for NDCG. This involved two crucial improvements over the regression based formulation of the simple NN:
\begin{itemize}
\item Moving to a pairwise preference formulation where the listings seen by a booker were used to construct pairs of \{booked listing, not-booked listing\} as training examples. During training we minimized cross entropy loss of the score difference between the booked listing over the not-booked listing.
\item Weighing each pairwise loss by the difference in NDCG resulting from swapping the positions of the two listings making up the pair. This prioritized the rank optimization of the booked listing towards the top of the search result list, instead of the bottom. For example, improving the rank of a booked listing from position $2$ to $1$ would get priority over moving a booked listing from position $10$ to $9$.
\end{itemize}
Table~\ref{table:tfcode} shows a partial implementation in TensorFlow\textsuperscript{TM}, in particular, how the pairwise loss was weighted.

\begin{table}
\begin{lstlisting}
def apply_discount(x):
    '''Apply positional discount curve'''
    return np.log(2.0)/np.log(2.0 + x)

def compute_weights(logit_op, session):
    '''Compute loss weights based on delta ndcg.
      logit_op is a [BATCH_SIZE, NUM_SAMPLES] shaped tensor
      corresponding to the output layer of the network.
      Each row corresponds to a search and each
      column a listing in the search result. Column 0 is the
      booked listing, while columns 1 through
      NUM_SAMPLES - 1 the not-booked listings. 
    '''
    logit_vals = session.run(logit_op)
    ranks = NUM_SAMPLES - 1 - logit_vals.argsort(axis=1).argsort(axis=1)
    discounted_non_booking = apply_discount(ranks[:, 1:])
    discounted_booking = apply_discount(np.expand_dims(ranks[:, 0], axis=1))
    discounted_weights = np.abs(discounted_booking - discounted_non_booking)
    return discounted_weight

# Compute the pairwise loss
pairwise_loss = tf.nn.sigmoid_cross_entropy_with_logits(
    targets=tf.ones_like(logit_op[:, 0]),
    logits=logit_op[:, 0] - logit_op[:, i:] )
# Compute the lambdarank weights based on delta ndcg
weights = compute_weights(logit_op, session)
#Multiply pairwise loss by lambdarank weights
loss = tf.reduce_mean(tf.multiply(pairwise_loss, weights))
\end{lstlisting}
\caption{TensorFlow\textsuperscript{TM} code to adapt pairwise loss to Lambdarank.}
\label{table:tfcode}
\end{table}

\subsection{Decision Tree/Factorization Machine NN}
\begin{figure}
\includegraphics[height=2in, width=3in]{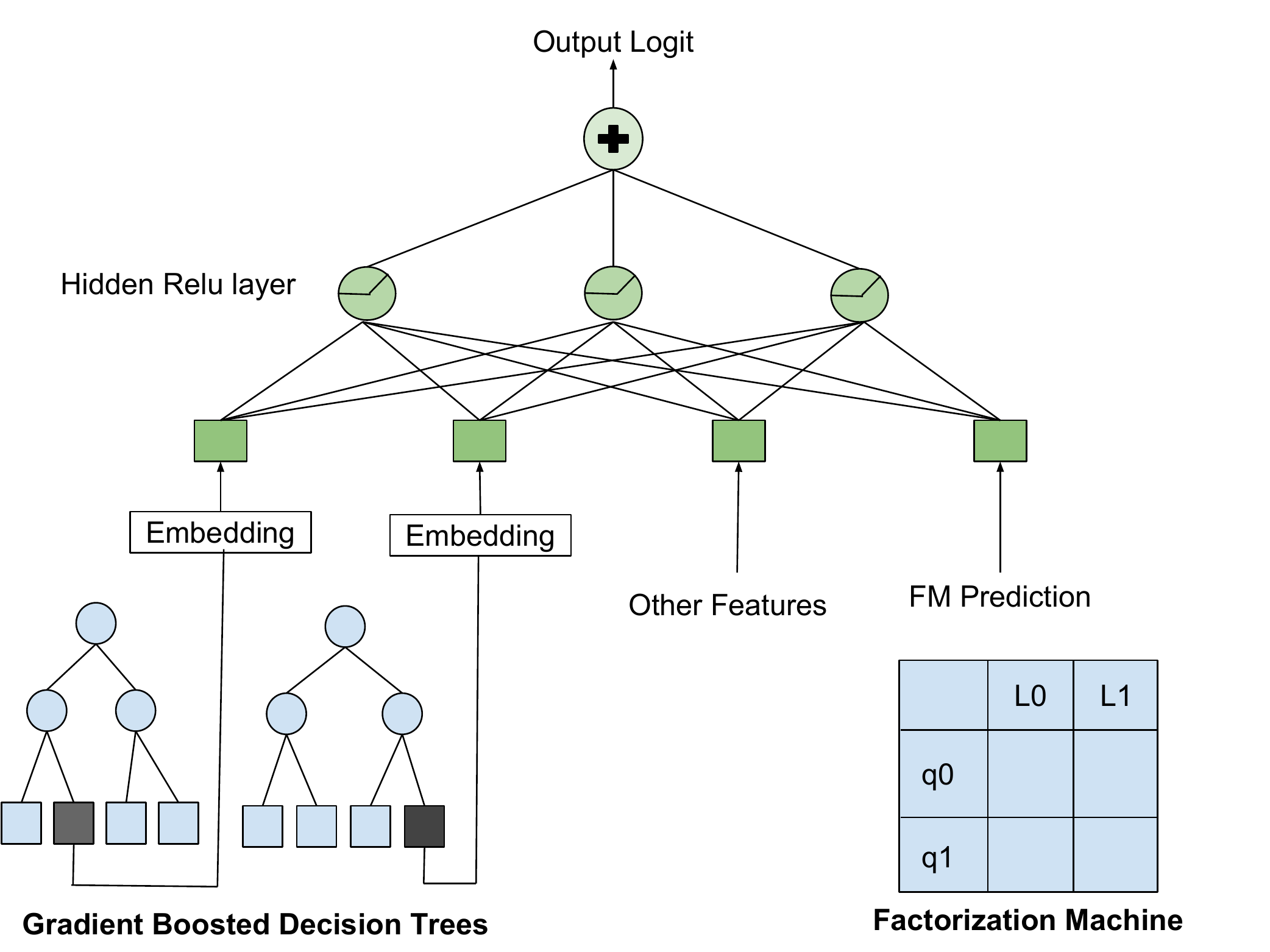}
\caption{NN with GBDT tree nodes and FM prediction as features}
\label{fig:gbdt}
\end{figure}
While the main ranking model serving production traffic was a NN at this point, we had other models under research. The notable ones were:
\begin{itemize}
 \item Iterations on the GBDT model with alternative ways to sample searches for constructing the training data.
 \item A factorization machine (FM)~\cite{fmachine} model that predicted the booking probability of a listing given a query, by mapping listings and queries to a $32$-dimensional space.
 \end{itemize}
 These new ways of looking at the search ranking problem revealed something interesting: although performances of the alternative models on test data were comparable to the NN, the listings upranked by them were quite different. Inspired by NN architectures like \cite{Wang}, the new model was an attempt to combine the strengths of all three models. For the FM model we took the final prediction as a feature into the NN. From the GBDT model, we took the index of the leaf node activated per tree as a categorical feature. Figure~\ref{fig:gbdt} gives an overview.

\subsection{Deep NN}
\begin{figure}
\includegraphics[height=2in, width=3in]{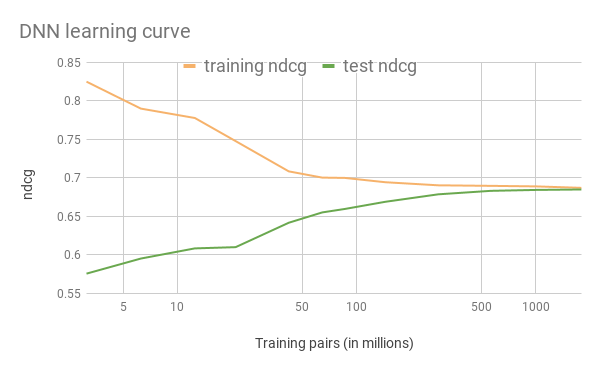}
\caption{DNN learning curve}
\label{fig:dnnlc}
\end{figure}
The complexity of the model at this point was staggering, and some of the issues mentioned in ~\cite{techdebt} begun to rear their heads. In our final leap, we were able to deprecate all that complexity by simply scaling the training data 10x and moving to a DNN with 2 hidden layers. Typical configuration of the network: an input layer with a total of $195$ features after expanding categorical features to embeddings, feeding the first hidden layer with $127$ fully connected ReLUs, and then the second hidden layer with $83$ fully connected ReLUs.

The features feeding the DNN were mostly simple properties of the listings such as price, amenities, historical booking count, etc, fed directly with minimal feature engineering. Exceptions include features output from another model: 
\begin{itemize}
\item Price of listings that have the Smart Pricing feature enabled, supplied by a specialized model~\cite{pricing}.
\item Similarity of the listing to the past views of the user, computed based on co-view embeddings~\cite{similarity}.
\end{itemize}
These models tap into data that isn't directly part of the search ranking training examples, providing the DNN with additional information.

To take a closer look at the DNN, we plot its learning curve in Figure~\ref{fig:dnnlc}, comparing NDCG over the training and test data set as a function of number of training pairs. Training on $1.7$ billion pairs, we were able to close the generalization gap between the training and the test data set.

Could we have launched the DNN directly, skipping all the stages of evolution? We try to answer this in the retrospective section, once more of the context surrounding the model is in place.

As a side note, while DNNs have achieved human level performance on certain image applications~\cite{imagenet}, it is very hard for us to judge where we stand for a similar comparison. Part of the problem is that it's unclear how to define human level performance. Going through the logs, it's quite challenging for us to identify which listing was booked. We find no objective truth in the logs, only tradeoffs highly conditional upon the budget and tastes of the guest which remain mostly hidden. Other researchers \cite{walmart} note the difficulty in using human evaluation even for familiar shopping items. For our application these difficulties are further exacerbated due to the novelty of the inventory.

Speaking of difficulties, next we discuss something rarely talked about: failed attempts.

\section{Failed Models}
The narrative of one successful launch followed by another presented in the previous section doesn't tell the whole story. Reality is studded with unsuccessful attempts that outnumber the successful ones. Retelling every attempt made would be time consuming, so we pick two particularly interesting ones. These models are interesting because they illustrate how some technique that is very effective and popular in the wild may not work as well when brought home.

\subsection{Listing ID}
Each listing at Airbnb has a corresponding unique id. One of the exciting new opportunities unlocked by NNs was to use these listing ids as features. The idea was to use the listing ids as index into an embedding, which allowed us to learn a vector representation per listing, encoding their unique properties. A reason for our excitement was the success other applications had achieved in mapping such high cardinality categorical features to embeddings, such as learning embeddings for words in NLP applications~\cite{deepnlp}, learning embeddings for video and user id in recommender systems~\cite{youtube}, etc.
\begin{figure}
\includegraphics[height=2in, width=3in]{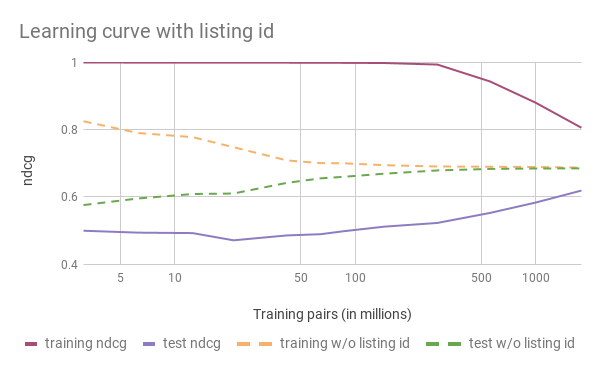}
\caption{Learning curve with listing id feature}
\label{fig:lilc}
\end{figure}

However, in the different variations we tried, listing ids mostly led to overfitting. Figure~\ref{fig:lilc} plots the learning curve from one such attempt, where we saw significant improvement in NDCG on the training set, but none on the test set.

The reason why such an established technique fails at Airbnb is because of some unique properties of the underlying marketplace. The embeddings need substantial amounts of data per item to converge to reasonable values. When items can be repeated without constraints, such as online videos or words in a language, there is no limit to the amount of user interaction an item can have. Obtaining large amounts of data for the items of interest is relatively easy. Listings, on the other hand, are subjected to constraints from the physical world. Even the most popular listing can be booked at most 365 times in an entire year. Typical bookings per listing are much fewer. This fundamental limitation generates data that is very sparse at the listing level. The overfitting is a direct fallout of this limitation.

\subsection{Multi-task learning}
\begin{figure}
\includegraphics[height=2in, width=3in]{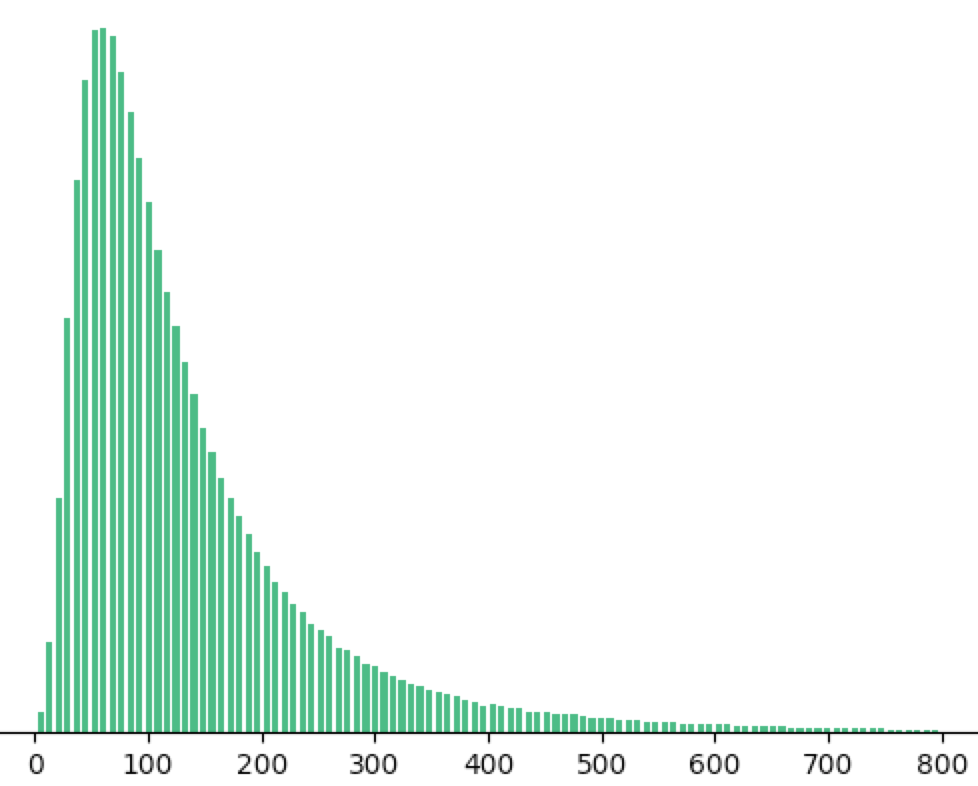}
\caption{Distribution of views to booking ratio across listings}
\label{fig:viewsperbooking}
\end{figure}

While bookings have a physical limitation, user views of the listing details pages are not constrained in the same way, and those we have in spades. Figure~\ref{fig:viewsperbooking} shows the distribution of views to bookings ratio for listings, with bookings typically orders of magnitude more sparse than views. Taking a step further, we found long views of listing details pages, unsurprisingly, correlated with bookings.
\begin{figure}
\includegraphics[height=3in, width=3.5in]{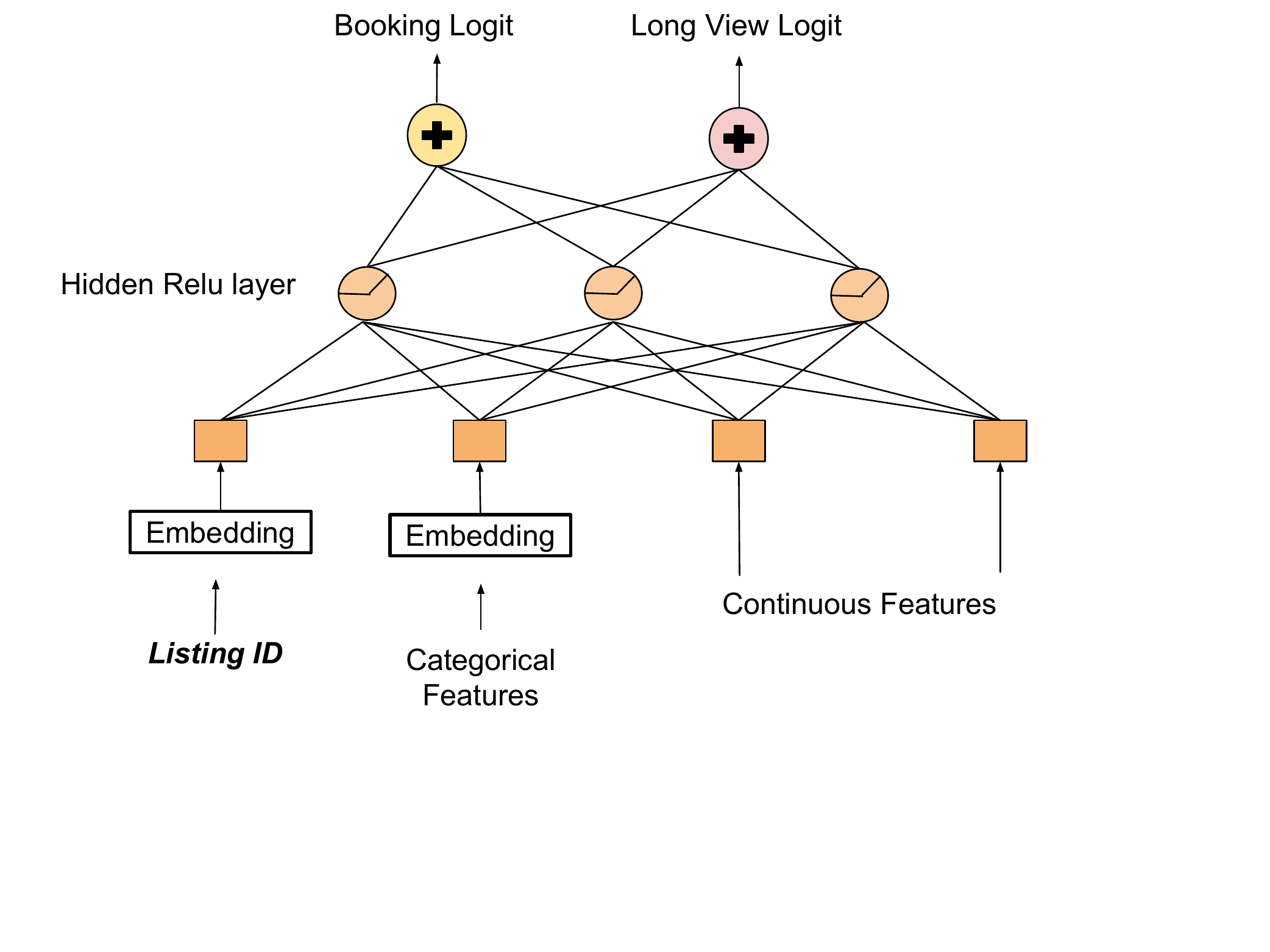}
\caption{Muti-task architecture predicting bookings and views}
\label{fig:multitask}
\end{figure}

To tackle the overfitting listing ids, we built a model taking a page out of the multi-task learning playbook~\cite{multitask}. The model simultaneously predicted the probability of booking and long view using two separate output layers; one optimized the loss with the booked listings as positive labels and the other targeted long views. Both output layers shared a common hidden layer as shown in Figure~\ref{fig:multitask}. Most importantly, the listing id embedding was shared as it was in the fan-in of the hidden layer. The idea was that the model would be able to transfer its learnings from long views to predict bookings and avoid overfitting. Since the number of long view labels outnumbered the booking labels by orders of magnitude, a compensating higher weight was applied to the booking loss to preserve focus on the booking objective. The loss for each long view label was further scaled by $log(view\_duration)$ as proposed in ~\cite{dwelltime}. For scoring listings online, we used the booking prediction only.

When tested online, the model increased long views by a large margin. But bookings remained neutral. Manually inspecting listings which had a high ratio of long views compared to bookings, we found several possible reasons which could have resulted in this gap. Such long views could be driven by high-end but high priced listings, listings with long descriptions that are difficult to parse, or extremely unique and sometimes humorous listings, among several other reasons. Once again, it is the uniqueness of the Airbnb marketplace where long views are correlated with bookings, but have a large orthogonal component as well that makes predicting bookings based on them challenging. A better understanding of listing views continues to be a topic of research for us.

\section{Feature Engineering}
The baseline GBDT pipeline we started with had extensive feature engineering. Typical transformations included computing ratios, averaging over windows, and other flavors of composition. The feature engineering tricks had accumulated over years of experimentation. Yet it was unclear if the features were the best they could be, or even up to date with the changing dynamics of the marketplace. A big attraction of NNs was to bring in feature automation, feeding raw data and letting the feature engineering happen in the hidden units of the NN driven by data.

Yet this section is dedicated to feature engineering, because we found that making NNs work efficiently involved a little more than feeding raw data. This flavor of feature engineering is different from the traditional one: instead of deriving the math to perform on the features before feeding them into the model, the focus shifts to ensuring the features comply with certain properties so that the NN can do the math effectively by itself.

\subsection{Feature normalization}
In our very first attempt at training a NN, we simply took all the features used to train the GBDT model and fed it to the NN. This went down very badly. The loss would saturate in the middle of training and additional steps would have no effect. We traced the issue to the fact that the features were not normalized properly.

For decision trees, the exact numeric values of the features hardly matter, as long as their relative ordering is meaningful. Neural networks on the other hand are quite sensitive to the numeric values the features take. Feeding values that are outside the usual range of features can cause large gradients to back propagate. This can permanently shut off activation functions like ReLU due to vanishing gradients~\cite{deadrelu}. To avoid it we ensure all features are restricted to a small range of values, with the bulk of the distribution in the \{-1, 1\} interval and the median mapped to 0. This by and large involves inspecting the features and applying either of the two transforms:
\begin{itemize}
\item In case the feature distribution resembles a normal distribution, we transform it by $(feature\_val - \mu)/\sigma$, where $\mu$ is the feature mean and $\sigma$ the standard deviation.
\item If the feature distribution looks closer to a power law distribution, we transform it by $log(\frac{1+feature\_val}{1+median})$.
\end{itemize}

\subsection{Feature distribution}
In addition to mapping features to a restricted numerical range, we ensured most of them had a smooth distribution. Why obsess over the smoothness of distributions? Below are some of our reasons.

\subsubsection{Spotting bugs}
When dealing with hundreds of millions of feature samples, how can we verify a small fraction of them are not buggy? Range checks are useful but limited. We found smoothness of distribution an invaluable tool to spot bugs as the distribution of errors often stand in contrast to the typical distribution. To give an example, we had bugs related to currencies in the prices logged for certain geographies. And for periods greater than 28 days, the logged price was the monthly price instead of the daily price. These errors showed up as spikes on the initial distribution plots.
 
\subsubsection{Facilitating generalization}
\begin{figure}
   \includegraphics[height=1.25in, width=1.75in]{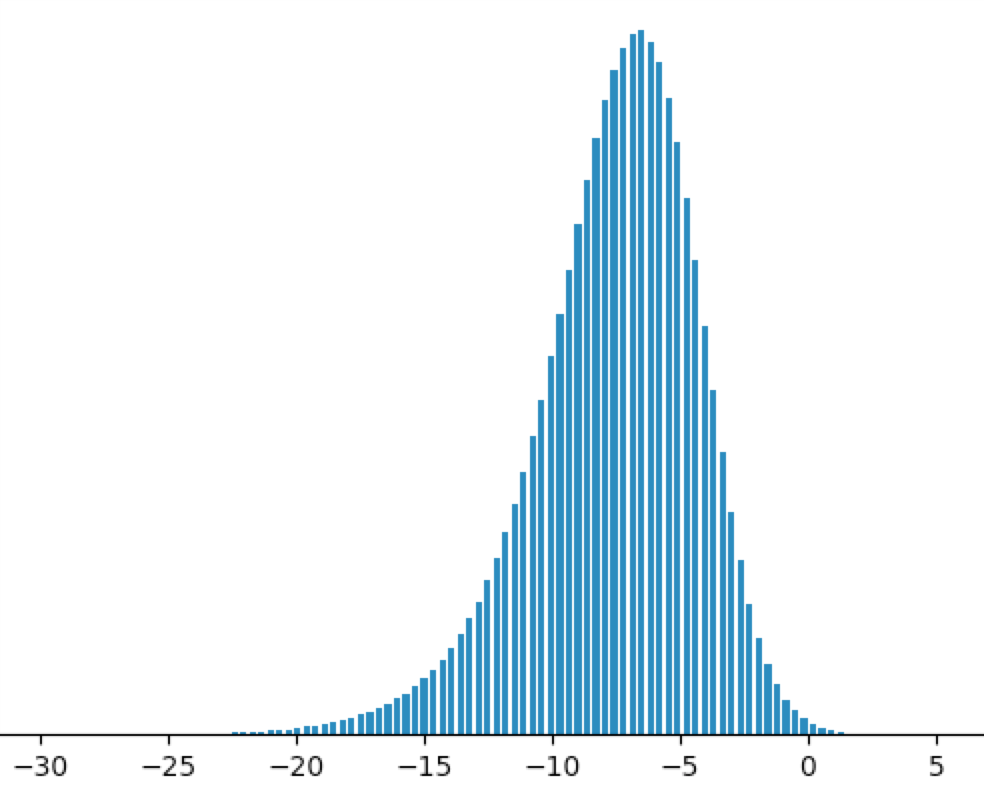}
   \caption{Distribution of output layer.}
   \label{fig:outputlayer}
\end{figure}
\begin{figure}
    \includegraphics[height=1in, width=1.5in]{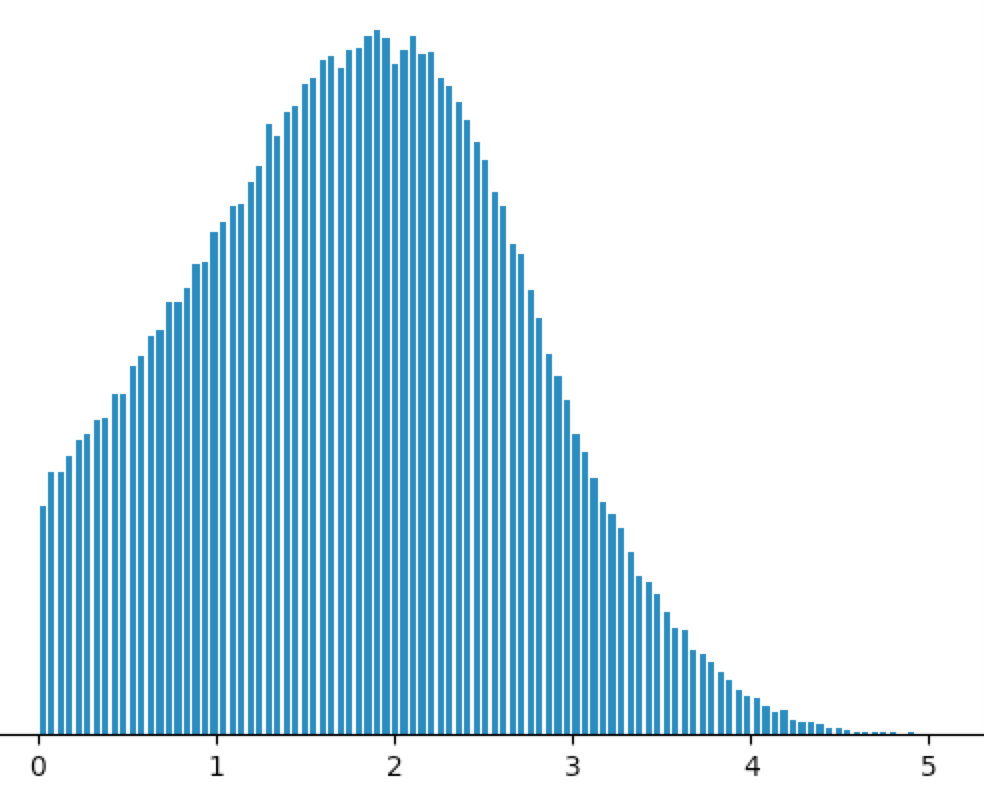}%
    \includegraphics[height=1in, width=1.5in]{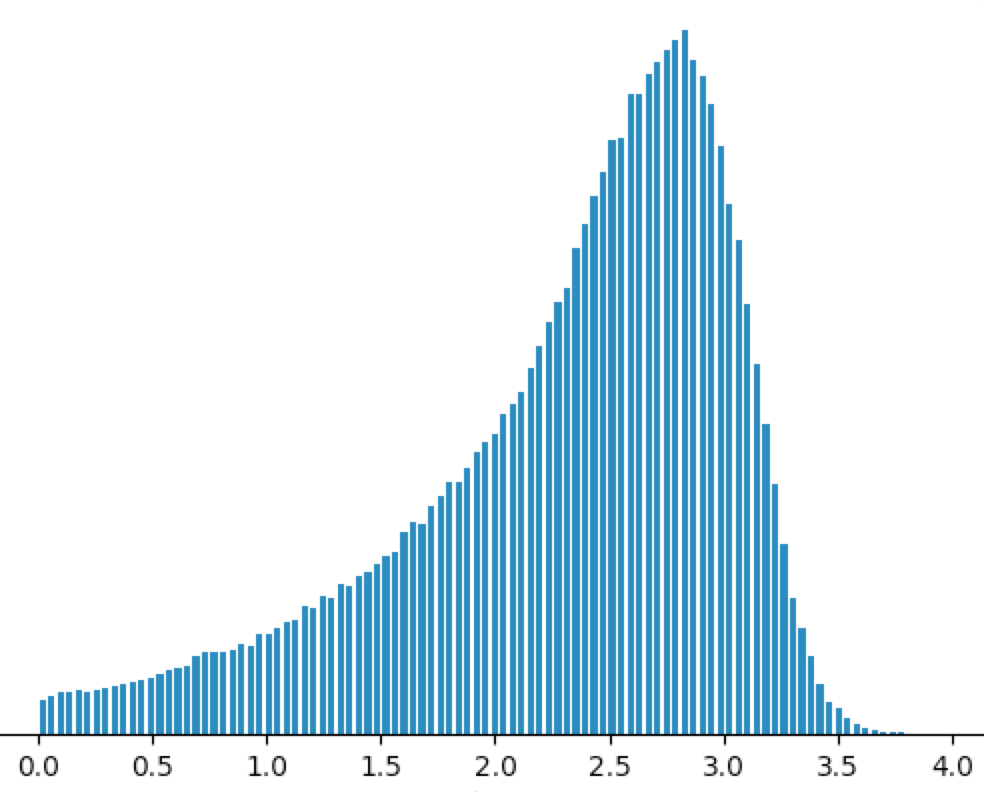}
    \caption{Example distributions from second hidden layer.}
    \label{fig:secondlayer}
\end{figure}
\begin{figure}
      \includegraphics[height=0.6in, width=1in]{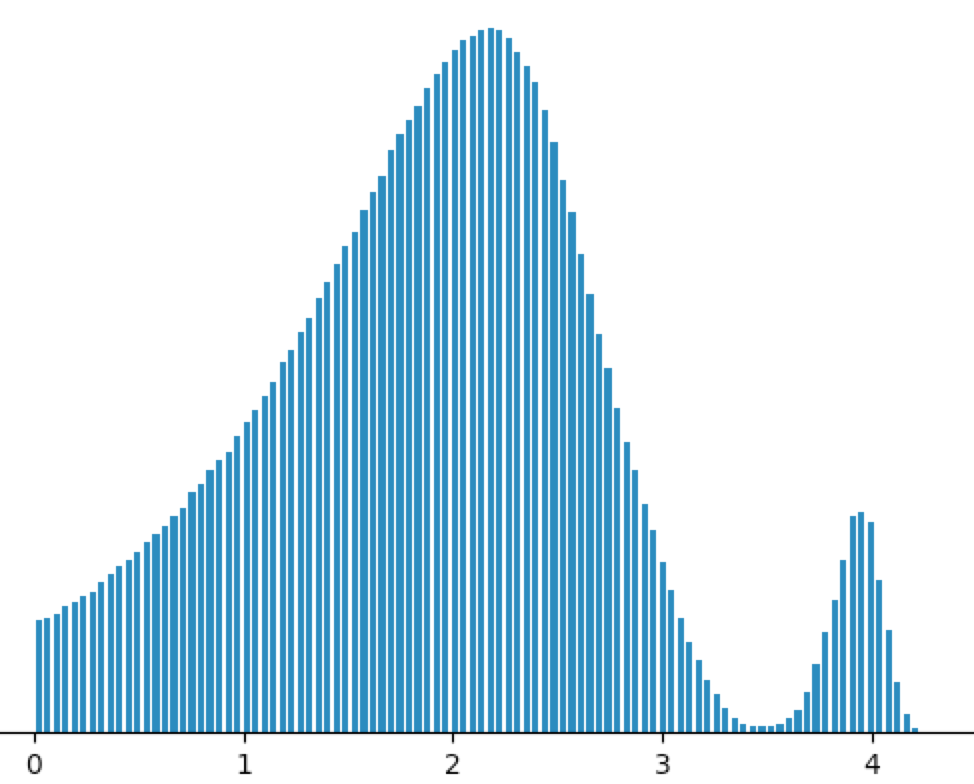}%
      \includegraphics[height=0.6in, width=1in]{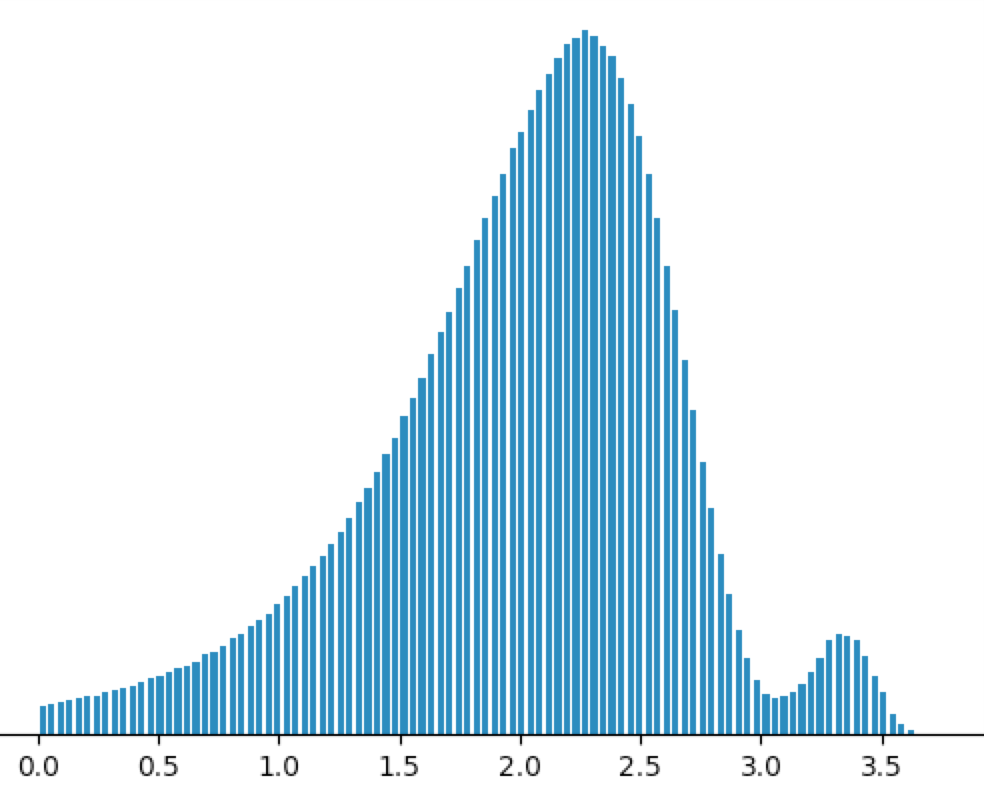}%
      \includegraphics[height=0.6in, width=1in]{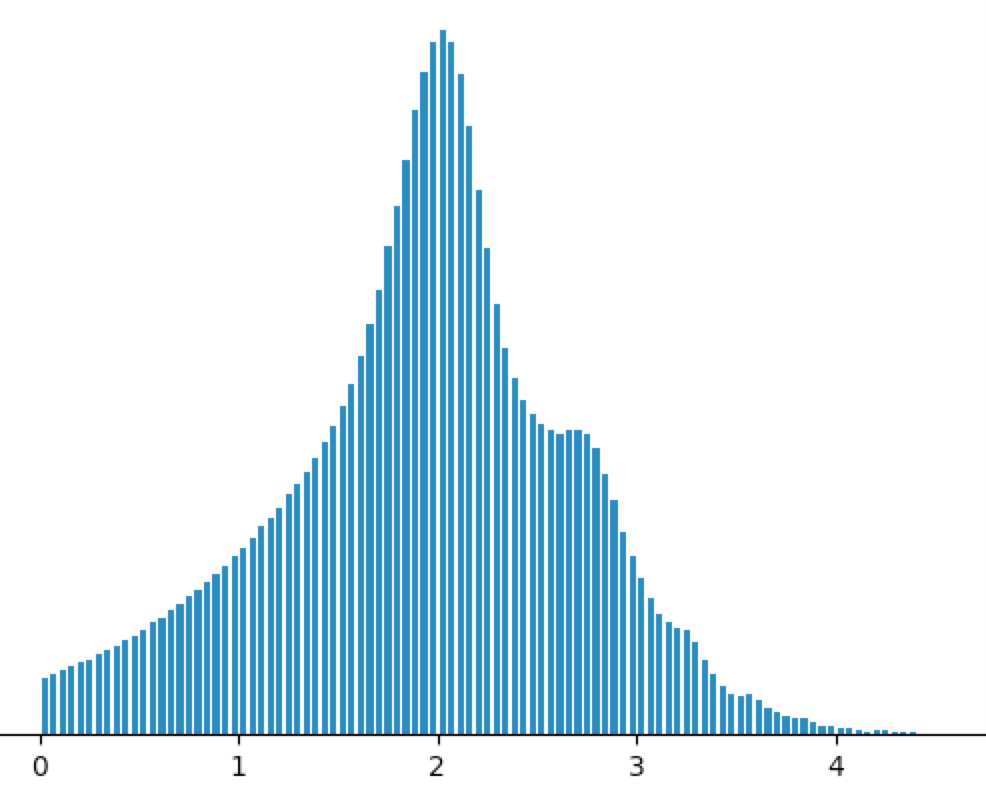}
   \caption{Example distributions from first hidden layer.}
   \label{fig:firstlayer}
\end{figure}
Answering exactly why DNNs are good at generalizing is a complicated topic at the forefront of research~\cite{generalization}. Meanwhile our working knowledge is based on the observation that in the DNNs built for our application, the outputs of the layers get progressively smoother in terms of their distributions. Figure~\ref{fig:outputlayer} shows the distribution from the final output layer, while Figure~\ref{fig:secondlayer} and Figure~\ref{fig:firstlayer} show some samples from the hidden layers. For plotting the values from the hidden layers, the zeros have been omitted and the transform $log(1 + relu\_output)$ applied. These plots drive our intuition for why DNNs may be generalizing well for our application. When building a model feeding on hundreds of features, the combinatorial space of all the feature values is impossibly large, and during training often a fraction of the feature combinations are covered. The smooth distributions coming from the lower layers ensure that the upper layers can correctly “interpolate” the behavior for unseen values. Extending this intuition all the way to the input layer, we put our best effort to ensure the input features have a smooth distribution.

How can we test if the model is generalizing well beyond logged examples? The real test is of course online performance of the model, but we found the following technique useful as a sanity check: scaling all the values of a given feature in the test set, such as price to 2x, 3x, 4x etc. and observing changes in NDCG. We found that the model's performance was remarkably stable over these values it had never seen before.

\begin{figure}
   \begin{subfigure}[b]{0.2\textwidth}
      \includegraphics[width=\textwidth]{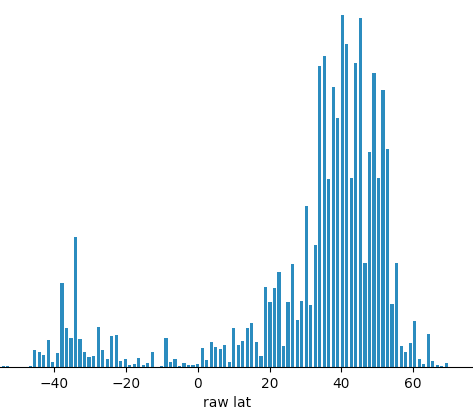}
      \caption{Distribution of raw lat}
      \label{fig:rawlat}
   \end{subfigure}
   \begin{subfigure}[b]{0.2\textwidth}
      \includegraphics[width=\textwidth]{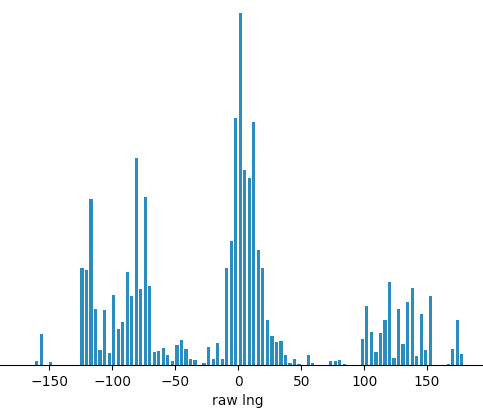}
      \caption{Distribution of raw lng}
       \label{fig:rawlng}
   \end{subfigure}
   \begin{subfigure}[b]{0.2\textwidth}
      \includegraphics[width=\textwidth]{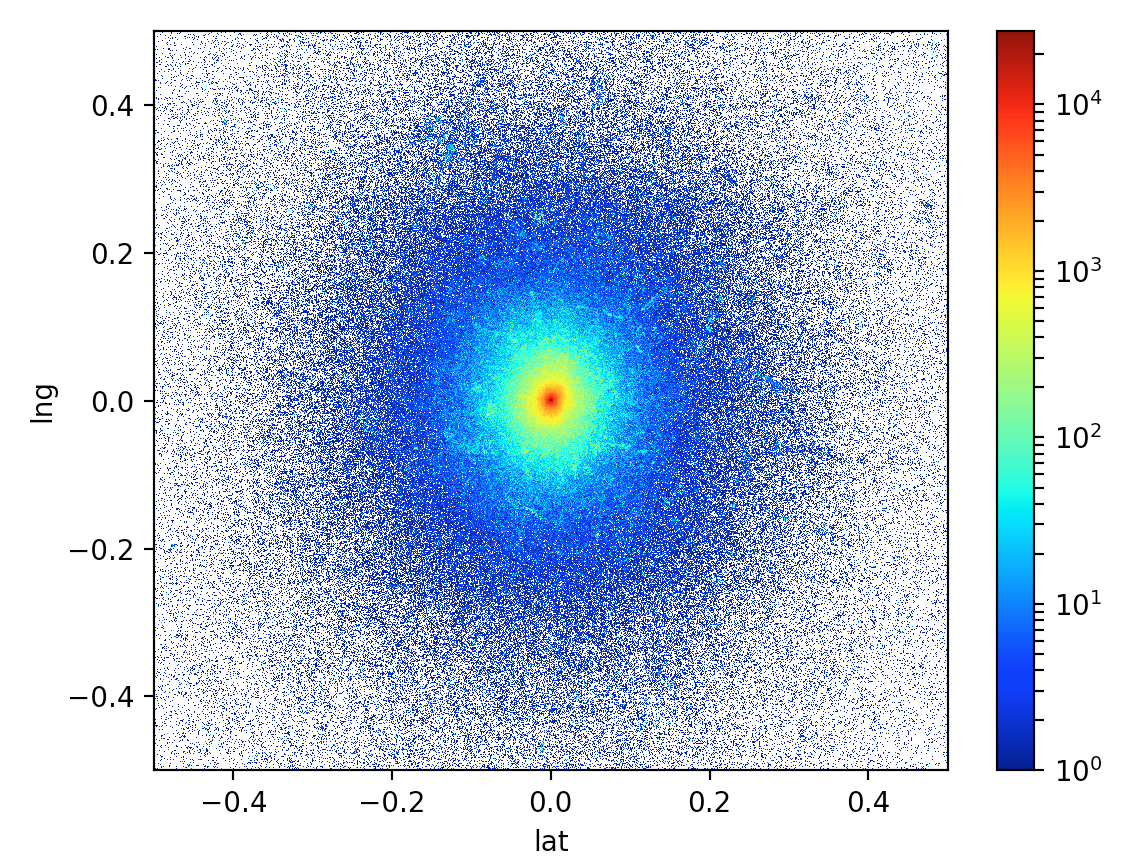}
      \caption{Heatmap of raw lat/lng offset}
      \label{fig:latlngoffset}
   \end{subfigure}
   \begin{subfigure}[b]{0.2\textwidth}
       \includegraphics[width=\textwidth]{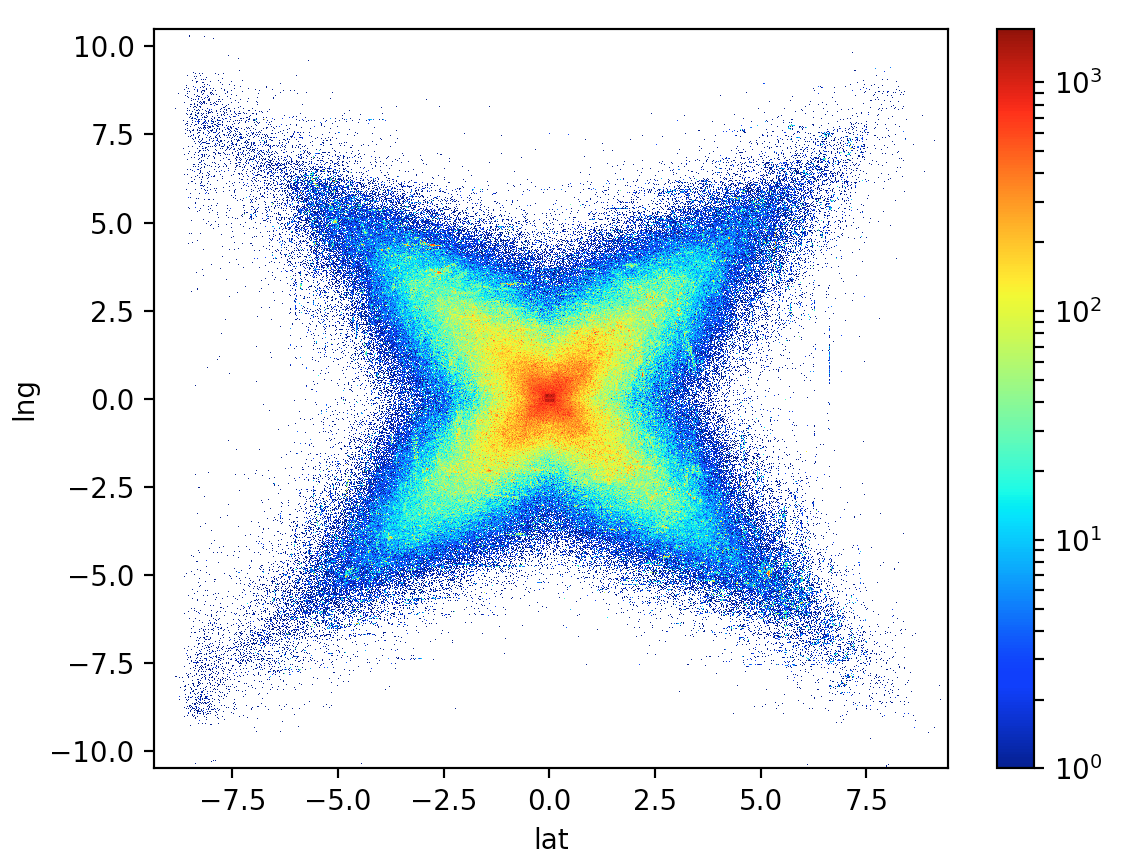}
       \caption{Heatmap of log lat/lng offset}
       \label{fig:loglatlngoffset}
   \end{subfigure}
   \begin{subfigure}[b]{0.2\textwidth}
       \includegraphics[width=\textwidth]{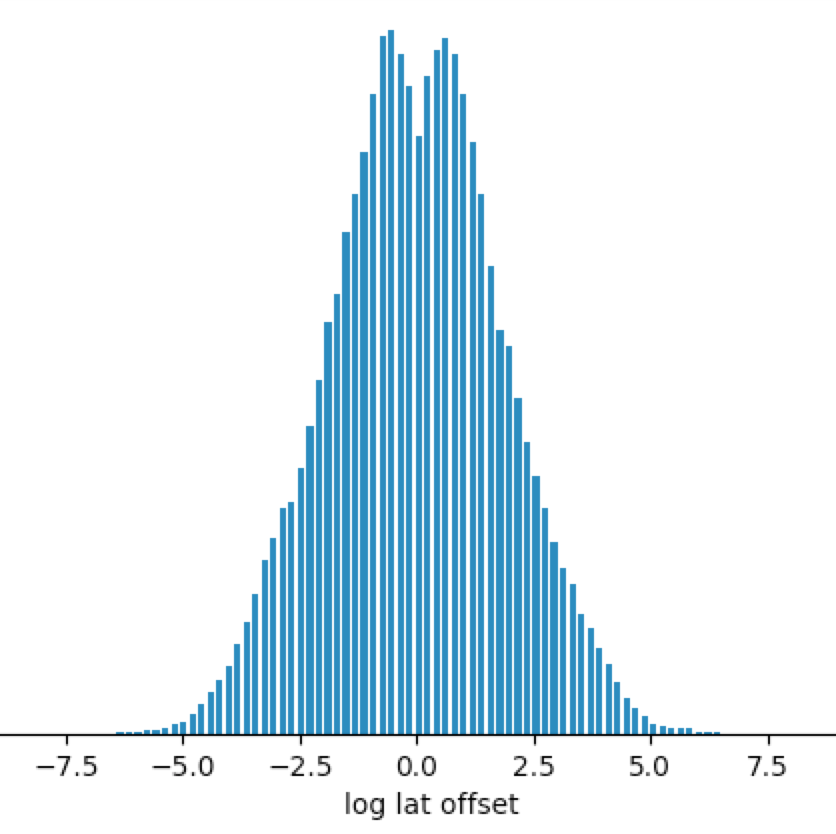}
      \caption{Distribution of log lat offset}
      \label{fig:loglat}
   \end{subfigure}
   \begin{subfigure}[b]{0.2\textwidth}
      \includegraphics[width=0.9\textwidth]{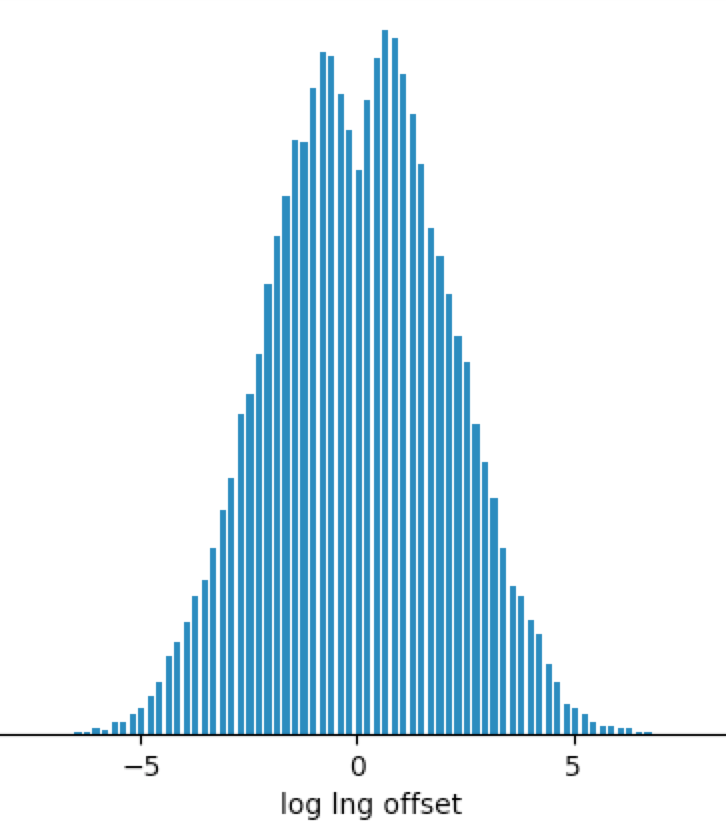}
      \caption{Distribution of log lng offset}
      \label{fig:loglng}
   \end{subfigure}
   \caption{Transforming geo location to smoothly distributed features}
   \label{fig:latlngfeature}
\end{figure}
Most features attained a smooth distribution once debugged and applied the appropriate normalization. However, for a few we had to do specialized feature engineering. An example is the geo location of a listing, represented by its latitude and longitude. Figure~\ref{fig:latlngfeature}\subref{fig:rawlat} and ~\ref{fig:latlngfeature}\subref{fig:rawlng} show the distribution of raw lat/lng. To make the distribution smooth, instead of raw lat/lng we compute the offset from the center of the map displayed to the user. Shown in Figure~\ref{fig:latlngfeature}\subref{fig:latlngoffset}, the mass seems concentrated at the center due to the tail end of maps which are zoomed out a lot. So we take $log()$ of the lat/lng offset, which yields the distribution in Figure~\ref{fig:latlngfeature}\subref{fig:loglatlngoffset}. This allows us to construct two features with smooth distributions, Figure~\ref{fig:latlngfeature}\subref{fig:loglat} and Figure~\ref{fig:latlngfeature}\subref{fig:loglng}.

To be clear, the raw lat/lng to offsets from map center transform is a lossy many-to-one function as it can convert multiple lat/lng to the same offset values. This allows the model to learn global properties based on distance rather than properties of specific geographies. To learn properties localized to specific geographies we use high cardinality categorical features that we describe later.

\subsubsection{Checking feature completeness}
\begin{figure}[!b]
   \begin{subfigure}[b]{0.20\textwidth}
      \includegraphics[width=\textwidth]{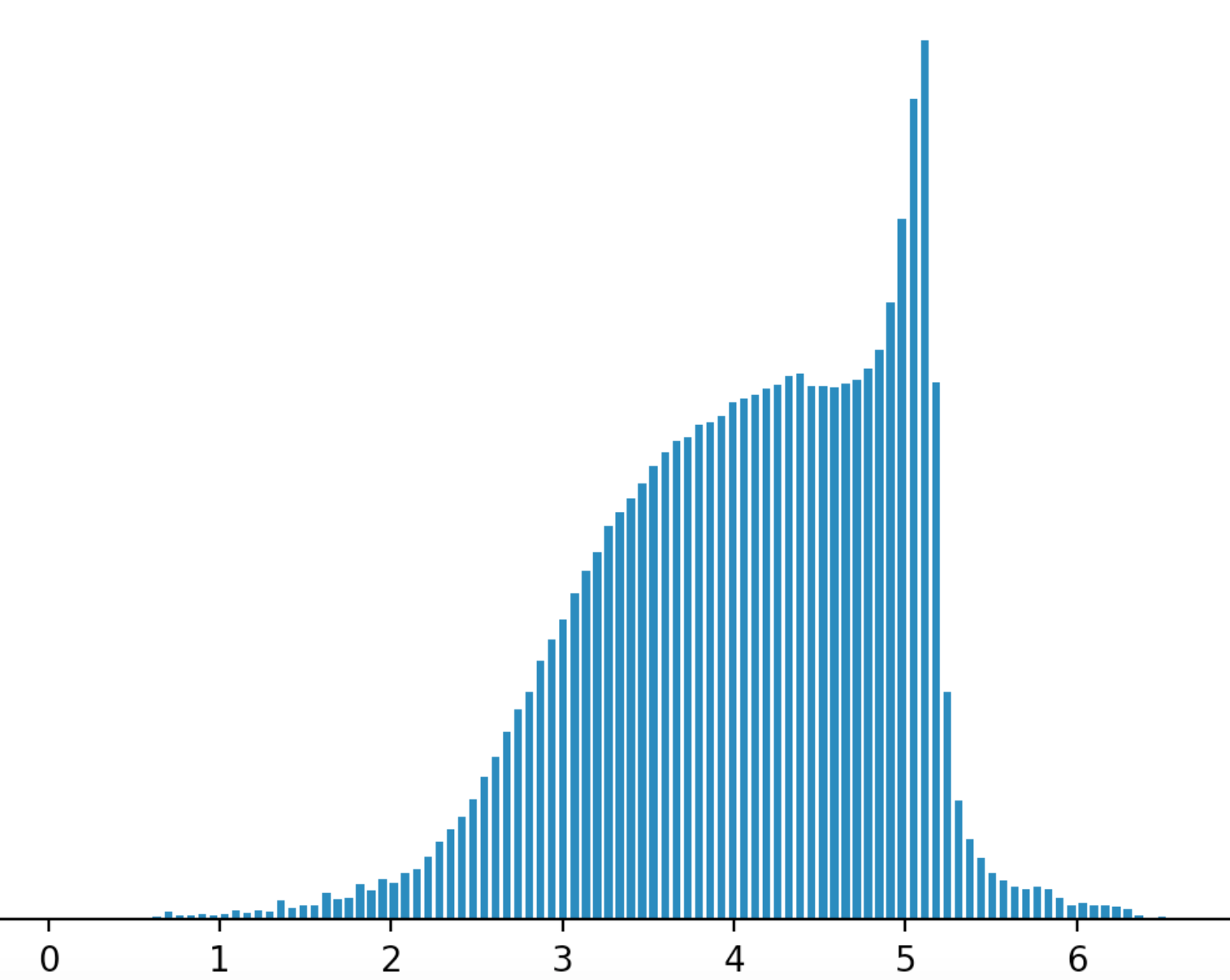}
      \caption{Distribution of raw occupancy}
      \label{fig:rawocc}
   \end{subfigure}
   \begin{subfigure}[b]{0.25\textwidth}
      \includegraphics[width=\textwidth]{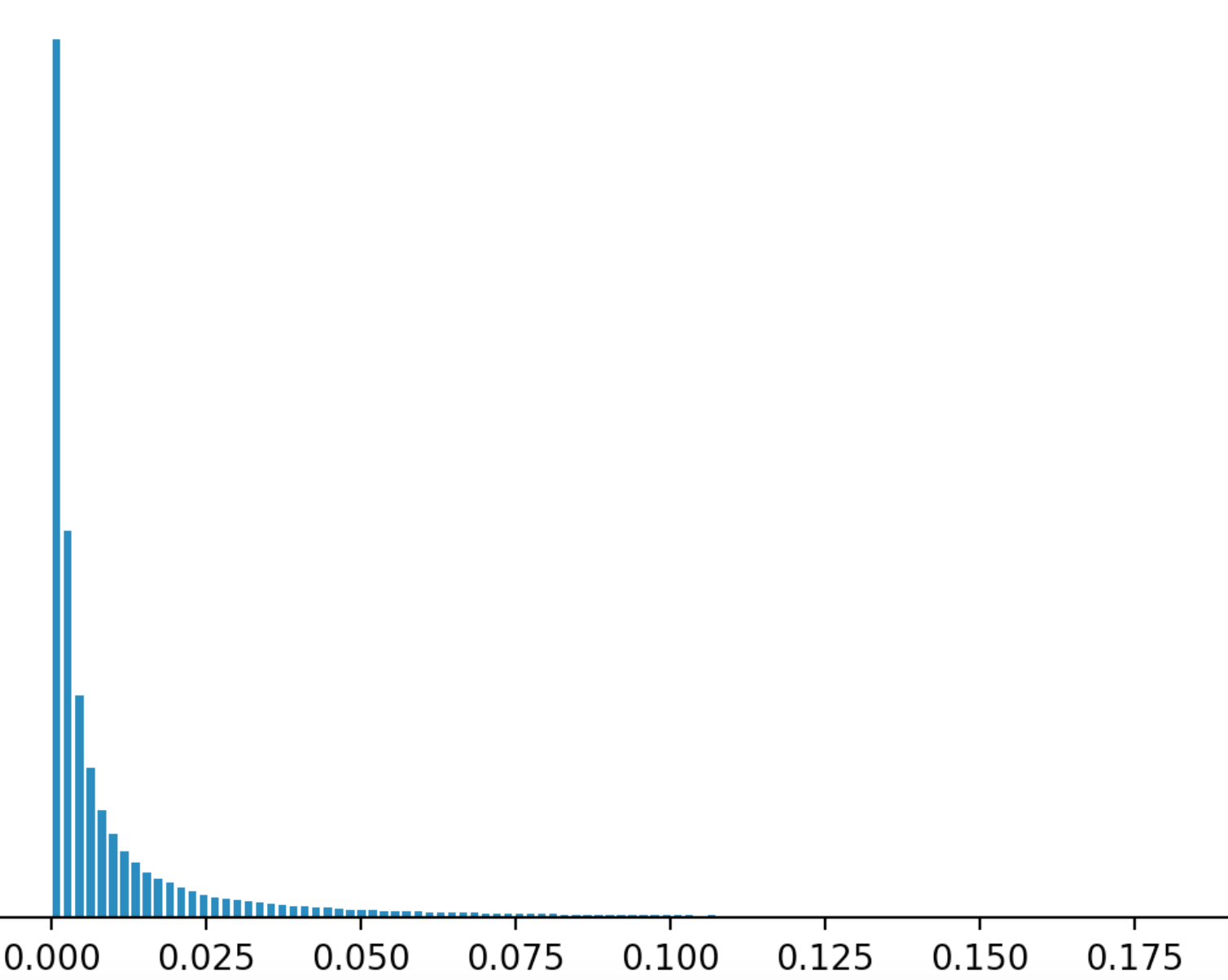}
      \caption{Distribution of occupancy normalized by average length of stay}
      \label{fig:normocc}
   \end{subfigure}
   \caption{}
   \label{fig:occ}
\end{figure}
In some cases, investigating the lack of smoothness of certain features lead to the discovery of features the model was missing. As an example, we had the fraction of available days a listing was occupied in the future as a signal of quality, the intuition being that high quality listings get sold out ahead of time. But the distribution of occupancy turned out to be perplexing in its lack of smoothness, shown in Figure~\ref{fig:occ}\subref{fig:rawocc}. After investigation we found an additional factor that influenced occupancy: listings had varying minimum stay requirements, sometimes extending to months, due to which they got occupied at different rates. However, we had not added the minimum required stay as a feature in the model as it was calendar dependent and considered too complex. But after looking at the occupancy distribution, we added average length of stay at the listing as a feature. Once occupancy is normalized by average length of stay, we see the distribution in Figure~\ref{fig:occ}\subref{fig:normocc}. Some features that lack a smooth distribution in one dimension may become smooth in a higher dimension. It was helpful for us to think through if those dimensions were already available to the model and if not, then adding them.

\subsection{High cardinality categorical features}
The overfitting listing id was not the only high cardinality categorical feature we tried. There were other attempts, where true to the promise of NNs, we got high returns with little feature engineering. This is best demonstrated by a concrete example. The preference of guests for various neighborhoods of a city is an important location signal. For the GBDT model, this information was fed by a heavily engineered pipeline, that tracked the hierarchical distribution of bookings over neighborhoods and cities. The effort involved in building and maintaining this pipeline was substantial. Yet it didn't factor in key elements like prices of the booked listings.

In the NN world, handling this information was simplicity itself. We created a new categorical feature by taking the city specified in the query and the level 12 S2 cell~\cite{s2cell} corresponding to a listing, then mapping the two together to an integer using a hashing function. For example, given the query "San Francisco" and a listing near the Embarcadero, we take the S2 cell the listing is situated in ($539058204$), and hash \{"San Francisco", $539058204$\} $\rightarrow 71829521$ to build our categorical feature. These categorical features are then mapped to an embedding, which feed the NN. During training, the model infers the embedding with back propagation which encodes the location preference for the neighborhood represented by the S2 cell, given the city query.

\begin{figure}
    \includegraphics[height=2.7861in, width=3.25in]{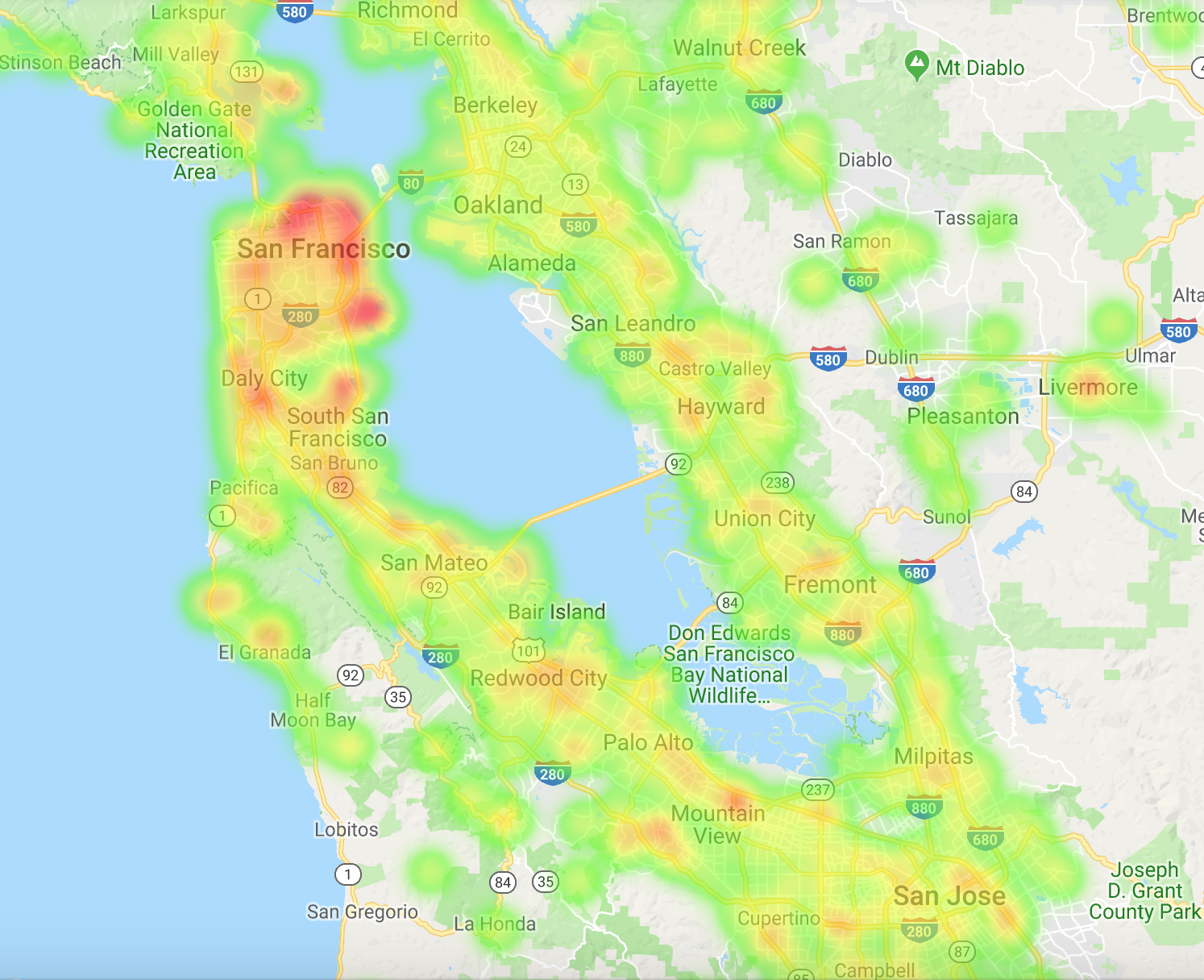}
    \caption{Location preference learnt for the query "San Francisco"}
    \label{fig:sfcross}
\end{figure}

Figure~\ref{fig:sfcross} visualizes the embedding values learnt for the query "San Francisco". This matched our own intuitive understanding of the area: the embedding values not only highlighted the major points of interest in the city, it indicated a preference for locations a little further south into the west bay instead of closer locations across the bridges that are major traffic snarls.

\section{System Engineering}
This section is concerned with speeding up training and scoring. A quick summary of our pipeline: search queries from a guest hits a Java\textsuperscript{TM} server that does retrieval and scoring. The server also produces logs which are stored as serialized Thrift\textsuperscript{TM} instances. The logs are processed using a Spark\textsuperscript{TM} pipeline to create training data. Model training is done using TensorFlow\textsuperscript{TM}. Various tools written in Scala and Java\textsuperscript{TM} are used for evaluating the models and computing offline metrics. The models are then uploaded to the Java\textsuperscript{TM} server that does retrieval and scoring. All these components run on AWS instances.

\paragraph{Protobufs and Datasets}
The GDBT model was fed training data in CSV format, and we reused much of this pipeline to feed the TensorFlow\textsuperscript{TM} models using feed\_dict. At first glance this seems like a “non-machine learning” issue, and was quite low in our list of priorities. Our wake up call came when we found our GPU utilizations near $\sim$25\%. Most of the training time was spent in parsing CSV and copying data through feed\_dict. We were effectively towing a Ferrari with a mule. Retooling the pipeline to produce training data as Protobufs and using Dataset~\cite{dataset} gave a 17x speedup to training and drove GPU utilization to $\sim$90\%. This ultimately allowed us to attack the problem by scaling the training data from weeks to months.

\paragraph{Refactoring static features}
A large number of our features were properties of listings that rarely changed. For instance, location, number of bedrooms, a long list of amenities, guest rules etc. Reading all these features as part of each training example created an input bottleneck. To eliminate this disk traffic, we used only the listing id as a categorical feature. All quasi-static listing features were packed as a non-trainable embedding indexed by the listing id. For the features that had a mutation over the training period, this traded off a small level of noise against training speed. Resident in the GPU memory, the embedding eliminated multiple kilobytes of data per training example that used to get loaded from disk via the CPU. This efficiency made it possible to explore a whole new class of models which took into account fine details about tens of listings the user had interacted with in the past.

\paragraph{Java\textsuperscript{TM} NN library}
Towards the beginning of 2017 when we started shipping the TensorFlow\textsuperscript{TM} models to production, we found no efficient solution to score the models within a Java\textsuperscript{TM} stack. Typically a back and forth conversion of the data between Java\textsuperscript{TM} and another language was required and the latency introduced in the process was a blocker for us. To adhere to the strict latency requirement of search, we created a custom neural network scoring library in Java\textsuperscript{TM}. While this has served us well till this point, we expect to revisit the issue to see the latest alternatives available.

\section{Hyperparameters}
While there were a few hyperparameters like number of trees, regularization, etc in the GBDT world, NNs take it to a whole new level. During initial iterations, we spent considerable time exploring this world mostly driven by anxiety akin to FOMO (fear-of-missing-out). The effort spent in surveying all the options and experimenting with the combinations didn't produce any meaningful improvements for us. However, the exercise did give us some confidence in our choices which we describe below.

\paragraph{Dropout}
Our initial impression was that dropout is the counterpart of regularization for neural networks~\cite{dropout}, and thereby essential. However for our application, the different flavors of dropout we tried, all lead to a slight degradation in offline metrics. To reconcile our lack of success with dropout, our current interpretation of it is closer to a data augmentation technique~\cite{dropout2}, effective when the randomness introduced mimic valid scenarios that may be missing in the training data. For our case, the randomness was simply producing invalid scenarios that was distracting the model.

As an alternative, we added hand crafted noise shapes taking into account the distribution of particular features, resulting in an improvement of $\sim$1\% in offline NDCG. But we failed to get any statistically significant improvement in online performance.

\paragraph{Initialization}
Out of sheer habit, we started our first model by initializing all weights and embeddings to zero, only to discover that is the worst way to start training a neural network. After surveying different techniques, our current choice is to use Xavier initialization~\cite{xavier} for the network weights and random uniform in the \{-1, 1\} range for embeddings.

\paragraph{Learning rate}
An overwhelming range of strategies confronted us here, but for our application we found it hard to improve upon the performance of Adam~\cite{adam} with its default settings. Currently we use a variant LazyAdamOptimizer~\cite{lazyadam}, which we found faster when training with large embeddings.

\paragraph{Batch size}
Varying batch size has dramatic effect on training speed, but its exact effect on the model itself is hard to grasp. The most useful pointer we found was ~\cite{learningrate}. We however, didn't quite follow the advice in the paper. Having swept the learning rate issue under the LazyAdamOptimizer carpet, we just opted for a fixed batch size of 200 which seemed to work for the current models.

\section{Feature Importance}
Estimating feature importance and model interpretability in general is an area where we took a step back with the move to NNs. Estimating feature importance is crucial in prioritizing engineering effort and guiding model iterations. The strength of NNs is in figuring out nonlinear interactions between the features. This is also the weakness when it comes to understanding what role a particular feature is playing as nonlinear interactions make it very difficult to study any feature in isolation. Next we recount some of our attempts in deciphering NNs.

\paragraph{Score Decomposition}
A homegrown partial dependence tool similar to ~\cite{partialdep} was the backbone of feature analysis in the GBDT world. In the NN world trying to understand individual feature importance only lead to confusion. Our first naive attempt was to take the final score produced by the network, and try to decompose it into contributions coming from each input node. After looking at the results, we realized the idea had a conceptual error: there was no clean way to separate the influence of a particular incoming node across a non-linear activation like ReLU.

\paragraph{Ablation Test}
This was another simplistic attack on the problem. The idea here was to ablate the features one at a time, retrain the model and observe the difference in performance. We could then assign the features importance proportional to the drop in performance their absence produced. However, the difficulty here was that any performance difference obtained by dropping a single feature resembled the typical noise in offline metrics observed anyway while retraining models. Possibly due to non-trivial redundancy in our feature set, the model seemed capable of making up for a single absent feature from the remaining ones. This leads to a Ship-of-Theseus paradox: can you keep ablating one feature at a time from the model, claiming it has no significant drop in performance?

\paragraph{Permutation Test}
We raised the sophistication in our next attempt, taking inspiration from permutation feature importance proposed for random forests~\cite{randomforests}. We observed the performance of the model on a test set after randomly permuting the values of a feature across the examples in the test. Our expectation was that more important a feature, the higher the resulting degradation from perturbing it. This exercise however lead to somewhat nonsensical results, like one of the most important features for predicting booking probability came out to be the number of rooms in a listing. The reason was that in permuting the features one at a time, we had baked in the assumption that the features were independent of each other, which was simply false. Number of rooms for instance is closely tied to price, number of guests staying, typical amenities etc. Permuting the feature independently created examples that never occurred in real life, and the importance of features in that invalid space sent us in the wrong direction. The test however was somewhat useful in determining features that were not pulling their weight. If randomly permuting a feature didn't affect the model performance at all, it was a good indication that the model was probably not dependent on it.

\paragraph{TopBot Analysis}
\begin{figure}
\includegraphics[height=1.5in, width=3.25in]{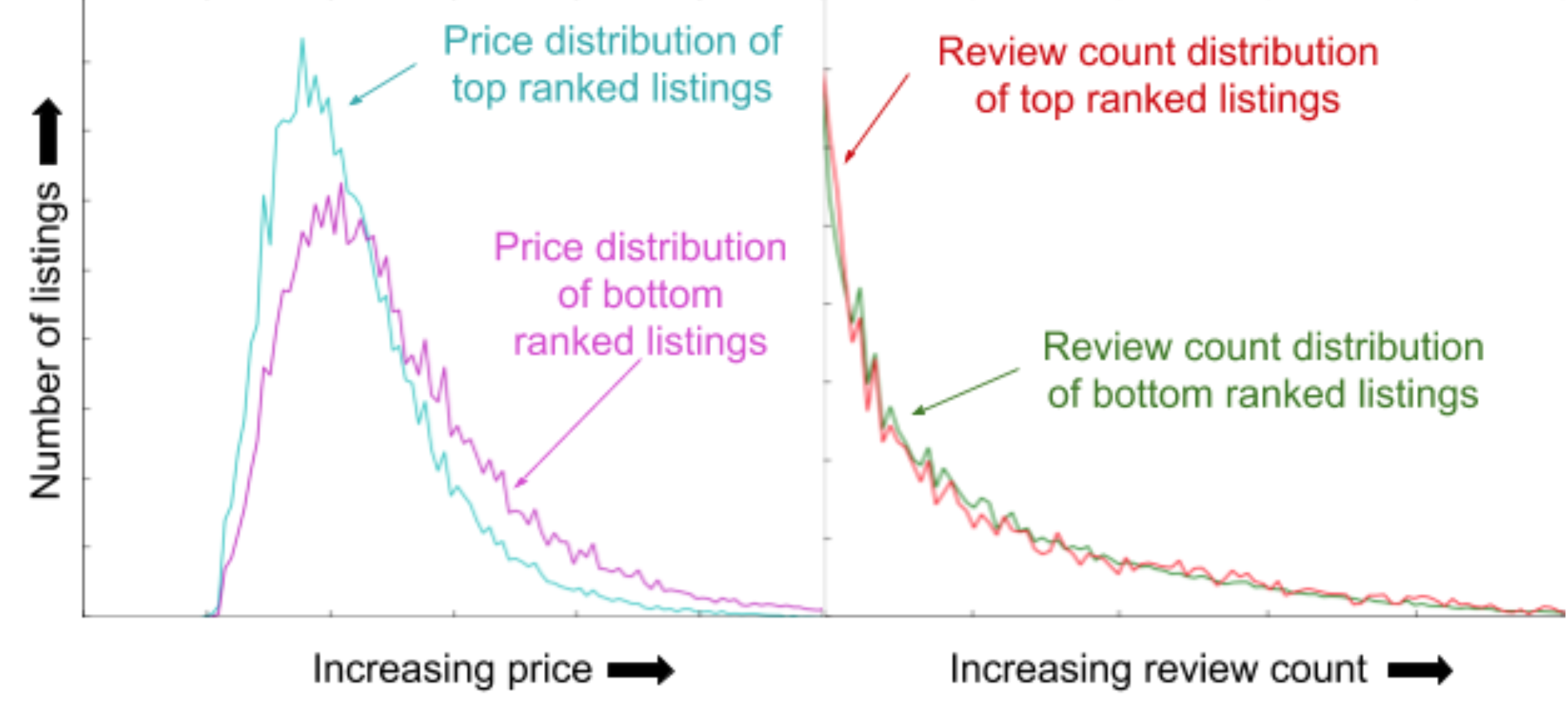}
\caption{Comparison of feature distribution for top and bottom ranked listings in test set.}
\label{fig:topbot}
\end{figure}
A homegrown tool designed to interpret the features without perturbing them in any way provided some interesting insights. Named TopBot, short for top-bottom analyzer, it took a test set as input and used the model to rank the listings per test query. It then generated distribution plots of feature values from the listings ranked at the top for each query, and compared them to the distribution of feature values from the listings at the bottom. The comparison indicated how the model was utilizing the features in the different value ranges. Figure~\ref{fig:topbot} shows an example. The distribution of prices for top ranked listings are skewed towards lower values, indicating the sensitivity of the model to price. However, the distribution of reviews look very similar when comparing top and bottom ranked listings indicating this version of the model was not utilizing reviews as expected, providing direction for further investigation. 

\section{Retrospective}
\begin{figure}
\includegraphics[height=1.5in, width=2.25in]{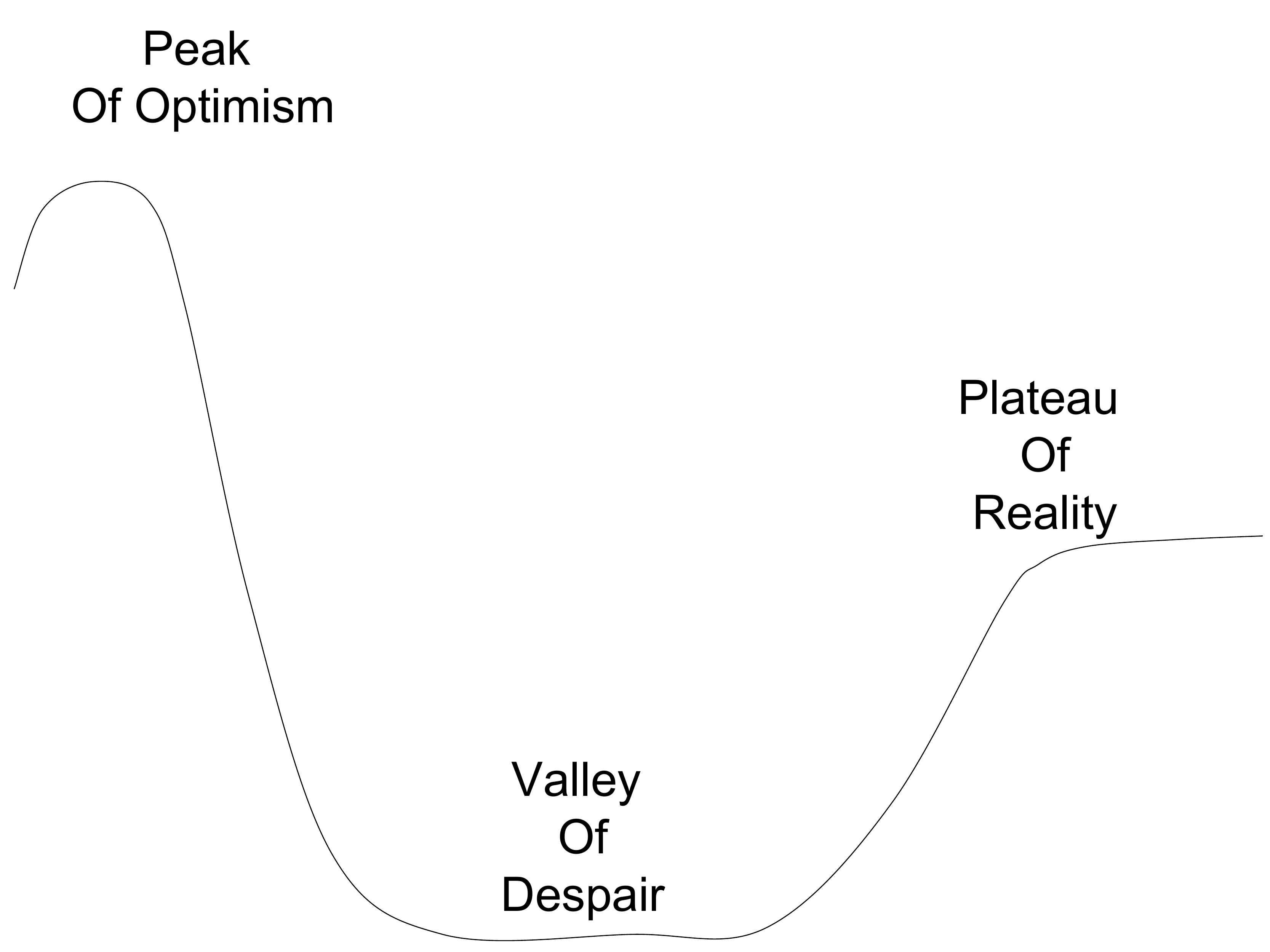}
\caption{Anatomy of a journey}
\label{fig:peakofoptimism}
\end{figure}
Figure~\ref{fig:peakofoptimism} summarizes our deep learning journey so far. Feeding on the ubiquitous deep learning success stories, we started at the peak of optimism, thinking deep learning would be a drop in replacement for the GBDT model and give us stupendous gains out of the box. A lot of initial discussions centered around keeping everything else invariant and replacing the current model with a neural network to see what gains we could get. This set us up for a plunge into the valley of despair, when initially none of those gains materialized. In fact, all we saw in the beginning was regression in offline metrics. Over time we realized that moving to deep learning is not a drop-in model replacement at all; rather it's about scaling the system. As a result, it required rethinking the entire system surrounding the model. Confined to smaller scales, models like GBDT are arguably at par in performance and easier to handle, and we continue to use them for focused medium sized problems.

So would we recommend deep learning to others? That would be a wholehearted Yes. And it's not only because of the strong gains in the online performance of the model. Part of it has to do with how deep learning has transformed our roadmap ahead. Earlier the focus was largely on feature engineering, but after the move to deep learning, trying to do better math on the features manually has lost its luster. This has freed us up to investigate problems at a higher level, like how can we improve our optimization objective, and are we accurately representing all our users? Two years after taking the first steps towards applying neural networks to search ranking, we feel we are just getting started.

\section{Acknowledgements}
Most of us have managed to bring down the metrics singlehandedly at some point. But lifting the metrics have always been the work of a collective. While naming everyone individually is not practical, we wish to thank those who directly worked towards making deep learning a success at Airbnb - Ajay Somani, Brad Hunter, Yangbo Zhu and Avneesh Saluja.

%\end{document}  % This is where a 'short' article might terminate